\documentclass{article}

\usepackage{microtype}
\usepackage{graphicx}
\usepackage{subfigure}
\usepackage{booktabs} 

\usepackage{hyperref}

\usepackage{amsmath}
\usepackage{caption}
\usepackage{subcaption}
\usepackage{wrapfig}
\usepackage{multirow}
\usepackage{amssymb}
\usepackage{pifont}
\usepackage{enumitem}

\usepackage[accepted]{icml2024}

\usepackage{amsmath}
\usepackage{amssymb}
\usepackage{mathtools}
\usepackage{amsthm}

\usepackage[capitalize,noabbrev]{cleveref}

\theoremstyle{plain}

\theoremstyle{definition}

\theoremstyle{remark}

\usepackage[textsize=tiny]{todonotes}

\icmltitlerunning{Generative Modeling of Quantum Distribution with Functional Flow Matching}

\begin{document}

\twocolumn[
\icmltitle{Generative Modeling of Quantum Distribution \\
            with Functional Flow Matching}



\icmlsetsymbol{equal}{*}

\begin{icmlauthorlist}
\icmlauthor{Jaehoon Hahm}{equal,yonsei1,gsds}
\icmlauthor{Tak Hur}{equal,yonsei1}
\icmlauthor{Joonseok Lee}{gsds,google}
\icmlauthor{Daniel K. Park}{yonsei1,yonsei2}
\end{icmlauthorlist}

\icmlaffiliation{gsds}{Graduate School of Data Science, Seoul National University, Seoul, Korea}
\icmlaffiliation{yonsei1}{Department of Statistics and Data Science, Yonsei University, Seoul, Korea}
\icmlaffiliation{yonsei2}{Department of Applied Statistics, Yonsei University, Seoul, Korea}
\icmlaffiliation{google}{Google Research, Mountain View, California, United States}

\icmlcorrespondingauthor{Daniel K. Park}{dkd.park@yonsei.ac.kr}

\icmlkeywords{Machine Learning, ICML}

\vskip 0.3in
]



\printAffiliationsAndNotice{\icmlEqualContribution} 

\begin{abstract}
The emergence of powerful deep generative models based on diffusion and flow matching has enabled the learning and modeling of complex distributions.
Learning quantum distributions, however, remains challenging due to the inherent difficulty of accurately modeling the meaningful physical properties of quantum states.
We propose Quantum Flow Matching (QFM), a novel generative model designed to learn quantum distribution by utilizing spin Wigner function and flow matching.
By converting density matrix into the spin Wigner function and leveraging functional flow matching to learn distributions in function space, QFM enables accurate and effective learning of multi-qubit quantum distributions.
We demonstrate the effectiveness of our method by evaluating physical quantities such as trace, purity, and entanglement entropy of the generated quantum states, accurately capturing the underlying physics of the given quantum distributions.
\end{abstract}

\section{Introduction}
\label{sec:intro}
Despite the unprecedented success of deep generative models, such as diffusion models \cite{song2020score, ho2020denoising} and flow matching \cite{lipman2022flow}, learning the distributions of quantum states remains a challenging task.
Leveraging state-of-the-art machine learning methods to model quantum states has been an interesting research direction \cite{carrasquilla2019reconstructing, carleo2019machine}.
However, applying modern generative models directly to learn quantum distributions has not been successful for several reasons.
First, existing diffusion and flow matching methods are solely focused on learning representations of classical data such as images or classical PDEs. These are inadequate for applications requiring consistency with important physical quantities such as purity, entanglement entropy, and quantum phases of matter.
Second, complex-valued density matrices pose a significant challenge due to the sign structure problem, making previous methods unsuitable for handling them \cite{westerhout2020generalization, dugan2023q}.


\begin{figure}[t]
    \centering
        \includegraphics[width=0.50\textwidth]{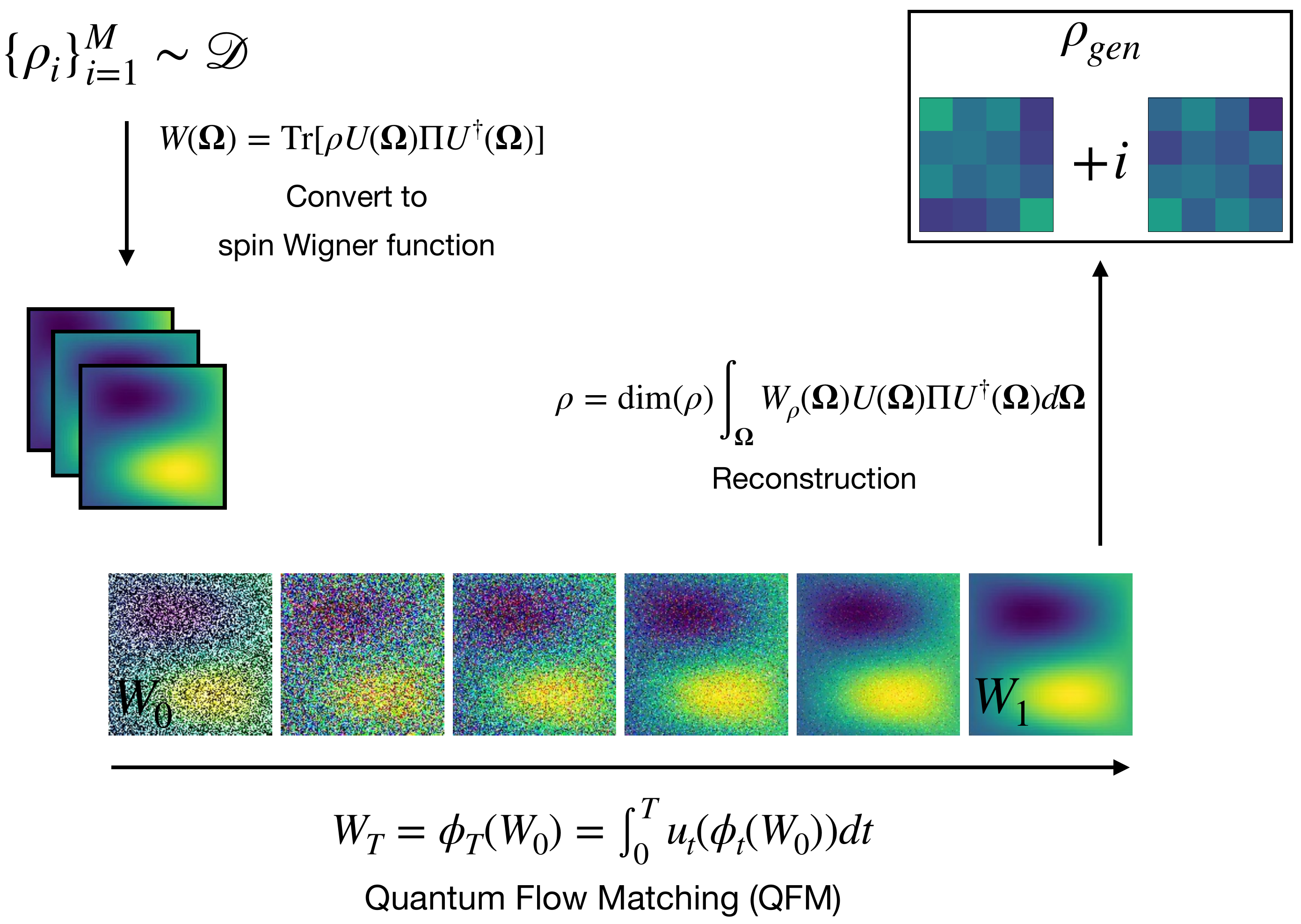}
        \vspace{-0.7cm}
        \caption{\textbf{Overall procedure of QFM}. We bypass the direct learning of quantum states by converting them into spin Wigner function and learn the underlying distribution. This approach allows us to effectively generate physically valid and accurate quantum states, which was not achievable by direct learning from density matrices.}
        \label{fig:QFM}
        \vspace{-0.5cm}
\end{figure}
To address this challenge, we propose Quantum Flow Matching (QFM), a novel generative modeling method that leverages flow matching to effectively learn the distributions of quantum states.
The procedure of QFM, summarized in Figure~\ref{fig:QFM}, can be illustrated in the following steps. 
First, we convert quantum states into informationally complete functional representations using spin Wigner functions.
Second, we employ Functional Flow Matching (FFM) to learn the distribution of spin Wigner functions in function space.
This allows us to generate new spin Wigner functions that accurately reflect the underlying quantum distribution. 
Finally, the new quantum states are reconstructed from these generated spin Wigner functions.

Our contributions are summarized as follows:
\begin{itemize}[leftmargin=1em,topsep=0pt,noitemsep]

  \item We propose QFM, a pioneering generative model for quantum distributions that accurately capture the underlying physics of a given quantum system.

  \item 
  We bypass the direct density matrix learning by using spin Wigner function. FFM is utilized to leverage resolution-invariant nature of functional models, enabling accurate reconstruction of quantum states from the learned spin Wigner functions.
  
  \item We demonstrate that our method effectively learns the underlying physics of given quantum systems by evaluating various physical quantities, such as trace, purity, and entanglement entropy.
  
\end{itemize}

\section{Learning Quantum Distribution with Flow Matching}
\label{sec:method}

\subsection{Quantum State to Spin Wigner Function}

\begin{figure}[t]
    \centering
        \includegraphics[width=0.42\textwidth]{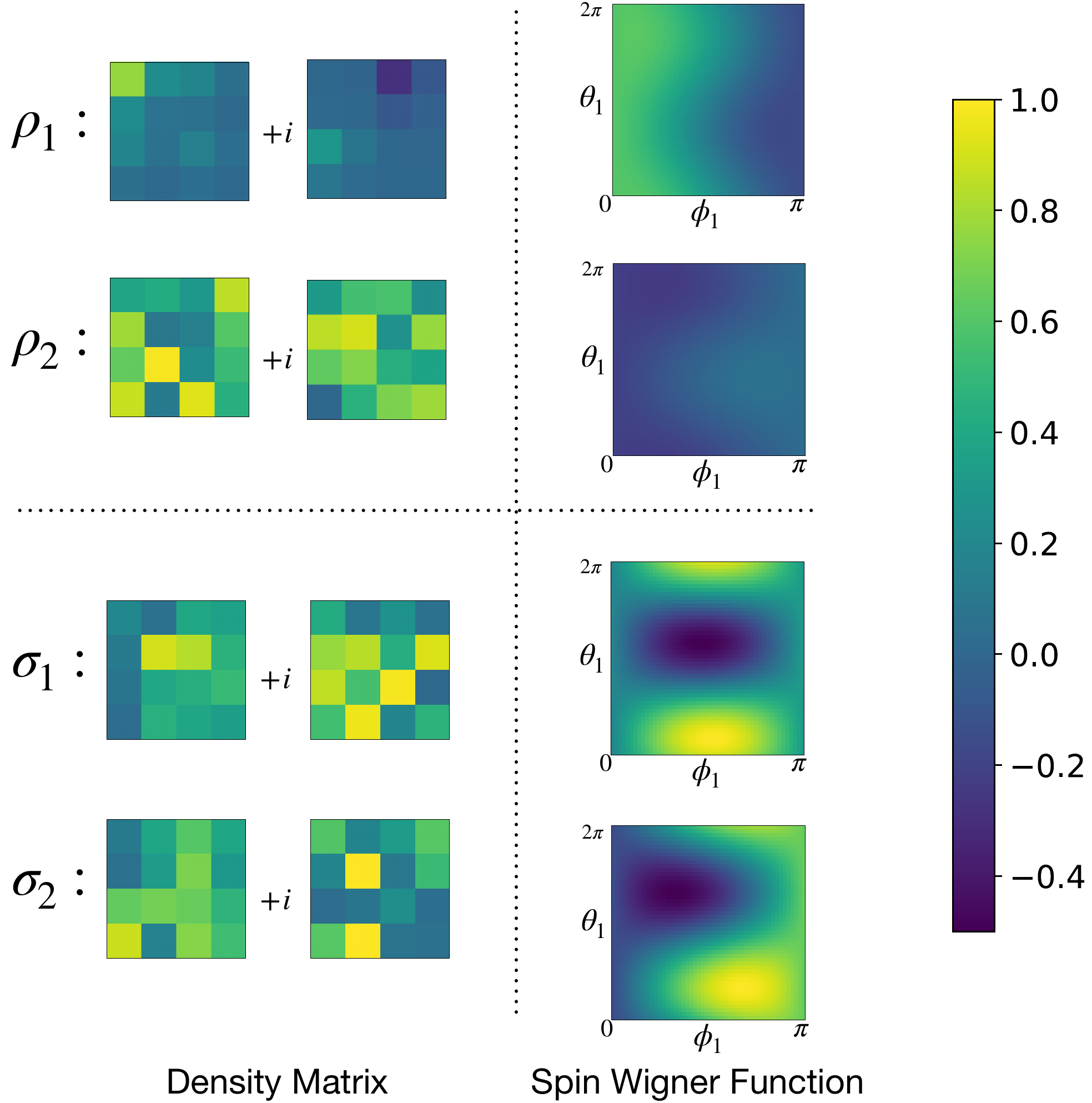}
        \caption{\textbf{An Example of two-qubit product states $\rho_1, \rho_2$ (top) and two-qubit maximally entangled states $\sigma_1, \sigma_2$ (bottom), in both density matrix (left) and spin Wigner function (right)}. 
        For the spin Winger functions, parameterizations on $\theta_1$  and $\phi_1$ are shown, with $\theta_2$ and $\phi_2$ fixed at $\pi$ and $\pi/2$, respectively. 
        Random arbitrary single-qubit rotation gates $U(\alpha, \beta, \gamma) = \exp(-i\alpha Z/2)\exp(-i\beta Y/2) \exp(-i\gamma Z/2)$ were applied to both qubits of $\vert 00 \rangle $ for the product states and to both qubits of $(\vert 00 \rangle + \vert 11 \rangle)/\sqrt{2}$ for the maximally entangled states. Unlike in the density matrix form, the product states and maximally entangled states are clearly distinguishable in the spin Wigner function representation.}
        \label{fig:spin Wigner}
        \vspace{-0.5cm}
\end{figure}

Learning the distributions of quantum states using deep generative models requires effective classical representations of these quantum states. 
Training the generative model using the state vector or density matrix representation alone is insufficient, as it does not ensure that important physical quantities are accurately captured.

The Wigner function effectively addresses these shortcomings by mapping a quantum density matrix onto a quasi-probability distribution in classical phase space~\cite{rundle2017simple, wigner1932quantum}.
However, since the Wigner function has been primarily explored in continuous variable and infinite-dimensional Hilbert spaces, such as in quantum optics and chemistry, its application in machine learning is still limited in these settings~\cite{dugan2023q}. 

In this work, we utilize spin Wigner function, which fully describes quantum systems of arbitrary (including finite) dimensions~\cite{tilma2016wigner, rundle2017simple}.
This approach enables the application of deep generative models to a range of intriguing tasks involving finite-dimensional Hilbert spaces, such as generating the ground state of a quantum spin chain with specific quantum phases.

For an arbitrary quantum state $\rho$, the spin Wigner function is given by $W(\mathbf{\Omega}) = \mathrm{Tr}[\rho U(\mathbf{\Omega}) \Pi U^\dagger(\mathbf{\Omega})]$, where $U(\mathbf{\Omega})$ and $\Pi$ are analogous to the displacement and parity operator in the original Wigner function, respectively. With an appropriate choice of Hermitian observable $\Pi$, along with unitary $U$ and parameterization $\mathbf{\Omega}$, the original density matrix can be fully reconstructed as $\rho = \mathrm{dim}(\rho) \int_{\mathbf{\Omega}} W(\mathbf{\Omega})U(\mathbf{\Omega}) \Pi U^\dagger(\mathbf{\Omega})d\mathbf{\Omega}$, where dim($\rho$) is the dimensionality of $\rho$.

Here, $d\mathbf{\Omega}$ should be chosen such that $U(\mathbf{\Omega})$ follows a Haar random, or unitary 2-design, distribution. Specifically, unitary 2-design, denoted by $\mathcal{U}$, is a distribution of unitary matrices where the expectation value matches that of the Haar distribution up to the second moment.
Then, the integral term of the reconstruction is given by
\begin{align}
    &\int_{\mathbf{\Omega}} W (\mathbf{\Omega}) \Delta(\mathbf{\Omega}) d\mathbf{\Omega} = \mathbb{E}_{\mathcal{U}} \mathrm{Tr}_A[(\rho \otimes I_B) U^{\otimes 2} \Pi^{\otimes 2} U^{\dagger\otimes 2}]  \nonumber
    \\
    &= \mathrm{Tr}_A[(\rho \otimes I_B )(c_{\Pi,I}I + c_{\Pi,F}F)] = c_{\Pi,I}I_B + c_{\Pi,F}\rho, \nonumber
\end{align}
where $\Delta(\mathbf{\Omega}) \equiv U(\mathbf{\Omega}) \Pi U^\dagger(\mathbf{\Omega})$, and expectation over $\mathcal{U}$ is taken with respect to the Haar distribution. $I$ and $F$ denote identity and swap operator on the composite system, respectively, and the constants are calculated as $c_{\Pi, I} = 0$ and $c_{\Pi, F} = 1/\mathrm{dim}(\rho)$~\cite{mele2024introduction}.
Therefore, we can fully reconstruct the original density matrix $\rho$.

To generalize this to $N$ qubits, there is flexibility in the choice of $U(\mathbf{\Omega})$ and $\Pi$. Throughout our work, we adhere to the conventions of~\citet{rundle2017simple}, where $\Pi_N$ is $2^N \times 2^N$ diagonal matrix with $2^{-N}[1 + (2^N-1)\sqrt{2^N + 1}]$ as the first element and $2^{-N}[1 - \sqrt{2^N + 1}]$ as the remaining elements.
This choice aligns with the previous calculations of $c_{\Pi,I}$ and $c_{\Pi, F}$.
The unitary operator and parameterization are generalized to $U_N(\boldsymbol{\Omega}) = \bigotimes_{j=1}^N e^{i \theta_j Z_j}e^{i \phi_j Y_j}$, where $\mathbf{\Omega} \equiv$ $(\vec\theta, \vec\phi)$ and $Z_j(Y_j)$ denotes the Pauli $Z(Y)$ operator on $j$-th qubit.

Specifically, we reconstruct $N$-qubit density matrix $\rho$ from spin Wigner Function by the following approximation:
 \begin{align}
     \rho &= \mathrm{dim}(\rho) \mathbb{E}_{\mathcal{U}} \left[W(\mathbf{\Omega})\Delta(\mathbf{\Omega}) \right] \nonumber \\
     &= \mathrm{dim}(\rho)   \int_{\bigotimes^N{S^2}} W(\mathbf{\Omega})\Delta(\mathbf{\Omega}) \prod_{i=1}^{N}{\sin{\phi_i}} d\theta_1 d\phi_1 \cdots d\theta_N d\phi_N  \nonumber \\
     &\simeq \pi^N \cdot \mathop{\mathbb{E}}_{\substack{\theta_i \sim U[0, 2\pi] \\ \phi_i \sim U[0, \pi]}} \left[W(\mathbf{\Omega})\Delta(\mathbf{\Omega}) \prod_{i=1}^{N}{\sin{\phi_i}} \right]. \nonumber 
\end{align}

Using this approach, we can construct an informationally complete representation of $N$-qubit density matrices with functions of $2N$ parameters.
Figure~\ref{fig:spin Wigner} demonstrates the two-qubit product states and maximally entangled states in the spin Wigner representation.

\subsection{Functional Flow Matching}

Flow Matching (FM) is a continuous normalizing flow that models the integration of vector field $u_t$, or the flow $\phi_t$, by learning the vector field of the prescribed dynamics.
This can be described with the ordinary differential equation $\frac{d}{dt} \phi_t(W_t) = u_t(\phi_t(W_t))$, where $W_t$ is the spin Wigner function at time $t \in [0,1]$.
\citet{lipman2022flow} proposed conditional flow matching objective (CFM), that learns the marginal vector field $u_t(W_t)$ which corresponds to the target probability density path $p_t(W_t)$, known to be more stable and robust compared to diffusion models \cite{ho2020denoising, song2020score}. The CFM objective is given by
\begin{align}
    \mathcal{L}_\text{CFM} = \mathbb{E}_{t \sim U(0,1), W_t \sim p_t(W_t)}{\|v_t(W_t) - u_t(W_t|W_1)\|^2}, \nonumber
\end{align}
where $v_t(W)$ is a trainable neural network.

To learn the distribution of spin Wigner function, we utilize Functional Flow Matching, which is a functional version of FM \cite{kerrigan2023functional}, generalized to the function space, implemented with Fourier Neural Operator \cite{li2020fourier}.
It features a resolution-invariant property of functional models, and we leverage it when reconstructing the generated spin Wigner functions back into the quantum states.
\section{Experiments}
\label{sec:experiments}

\textbf{Single Qubit: Trace, Purity.}
We first demonstrate the generation of single-qubit density matrices using QFM.
Our objective is to generate quantum states with a specified purity $\alpha = \mathrm{Tr}[\rho^2]$, where $\rho$ denotes density matrix of the quantum state.
Moreover, we also examine the trace of the generated data to ensure it satisfies the unit trace constraint.

We conduct experiments with a synthetic dataset consisting of 7000 pure states ($\alpha = 1$) and mixed states ($ \alpha \in \lbrace 0.625, 0.905 \rbrace$).
Single-qubit quantum states can be expressed as $(I + \mathbf{n} \cdot \boldsymbol{\sigma})/2$, where $\mathbf{n}$ and $\boldsymbol{\sigma}$ are Bloch and Pauli vectors, respectively. Datasets of purity $\alpha$ are prepared by sampling $\mathbf{n}$ from uniform distribution with the constraint $\| \mathbf{n} \|_2 = \sqrt{2 \alpha -1}$.



\begin{table}
    \vspace{-0.3cm}
    \caption{\textbf{Quantitative comparisons of QFM and direct learning of FM on a single qubit.} $\alpha$, $\mathrm{Tr}[\rho]$ and $\mathrm{Tr}[\rho^2]$ represent the purity of training dataset, average values of the trace, and the purity over the generated quantum states, respectively.}
    \centering
    \small
    \begin{tabular}{lc|c|c} 
        \toprule
            Method & $\alpha$ & $\mathrm{Tr}[\rho]$ & $\mathrm{Tr}[\rho^2]$ \\ 
        \midrule
            FM & 1 &  0.998  & 0.714  \\
            QFM & 1 &  0.998  & \textbf{1.004} \\
        \midrule
            FM & 0.625  &  1.001  & 0.531  \\
            QFM & 0.625 & 0.997  & \textbf{0.625}  \\
        \midrule    
            FM & 0.905  &  0.999  & 0.622  \\
            QFM & 0.905 & 0.995 & \textbf{0.895}  \\
        
        \bottomrule    
    \end{tabular}
    \label{table:quant_single}
    \vspace{-0.5cm}
\end{table}

\begin{table}
    \caption{\textbf{Quantitative comparisons of QFM and direct learning of FM on multi qubits.} $S$, $\mathrm{Tr}[\rho]$ and $S(\rho_A)$ represent the entanglement entropy of training dataset, the average values of the trace, and the entanglement entropy over the generated quantum states, respectively.}
    \centering
    \small

    \centering
    \begin{tabular}{lc|c|c} 
        \toprule
            Method & $S$ & $\mathrm{Tr}[\rho]$ & $\mathcal{S}(\rho_A)$ \\ 
        \midrule
            FM  & 0.7 & 1.009   & 0.783   \\
            QFM & 0.7 & 0.934   & \textbf{0.694} \\
        \bottomrule    
    \end{tabular}
        
    \label{table:quant_multi}
    \vspace{-0.5cm}
\end{table}

\textbf{Multi qubits: Trace, Entanglement Entropy.}
We further demonstrate the capability of QFM in a multi-qubit setting.
Specifically, we focus on generating 2-qubit quantum states with a desired entanglement entropy $S = - \mathrm{Tr}[\rho_A \mathrm{log} \rho_A]$, where $\rho_A$ is the reduced density matrix.
We prepare a synthetic dataset consisting of 900 quantums states with fixed entanglement entropy $S=0.7$. This is done by initializing two-qubit quantum states with an entanglement entropy of 0.7 and applying random arbitrary single-qubit rotations.

Tables~\ref{table:quant_single} and~\ref{table:quant_multi} present a comparative analysis between QFM and direct learning of FM using density matrices. QFM effectively captures the physical quantities of the dataset compared to direct learning of FM. This is demonstrated by its better performance in the target purity values (Table~\ref{table:quant_single}) and entanglement entropy (Table~\ref{table:quant_multi}).

\section{Future Work}
\label{sec:conclusion}

One potential future direction is extending the method to larger multi-qubits system such as generating genuinely multipartite entangled states \cite{goyeneche2014genuinely} and the ground states of generalized cluster Hamiltonian with specific quantum phases~\cite{caro2022generalization, gil2024understanding}.
Another interesting direction is quantum error mitigation by leveraging generative model as denoiser and formulating as solving an inverse problem \cite{kawar2022denoising, chung2022improving, shim2024diffusion}. 
Lastly, by simultaneously utilizing Quantum Autoencoders \cite{romero2017quantum}, our method potentially enables the accurate generation of high-dimensional quantum states, which have been intractable due to heavy computational costs.




\bibliography{reference}
\bibliographystyle{icml2024}




\end{document}